\documentclass{article}
\usepackage{spconf,amsmath,graphicx}
\usepackage{multirow}
\usepackage{booktabs}
\usepackage{amsmath}
\usepackage{bbm}
\usepackage{enumitem}
\usepackage{color}

\definecolor{my_yellow}{RGB}{191,144,0}
\definecolor{my_green}{RGB}{26,158,157}
\definecolor{my_blue}{RGB}{68,114,196}

\title{Inductive Relation Prediction from Relational Paths and Context\\ with Hierarchical Transformers}
\name{Author(s) Name(s)\thanks{Thanks to XYZ agency for funding.}}
\address{Author Affiliation(s)}
\name{Jiaang Li$^{\dagger}$, \qquad Quan Wang$^{\ddagger\ast}$, \qquad Zhendong Mao$^{\dagger}$
\thanks{$\ast$ Corresponding author. This work is supported
by the National Key Research and Development Program of China (No.2022YFC3302101), and the National Natural Science Foundation of China, Grant No.62222212, No.62232006, No.61876223 }
}
\address{$^{\dagger}$ University of Science and Technology of China, Hefei, China \\
$^{\ddagger}$Beijing University of Posts and Telecommunications, Beijing, China\\
{\small \tt jali@mail.ustc.edu.cn, wangquan@bupt.edu.cn, zdmao@ustc.edu.cn}
}

\begin{document}
\ninept
\maketitle
\begin{abstract}
Relation prediction on knowledge graphs (KGs) is a key research topic.
Dominant embedding-based methods mainly focus on the \textit{transductive} setting and lack the \textit{inductive} ability to generalize to new entities for inference.
Existing methods for inductive reasoning mostly mine the connections between entities, i.e., relational paths,
without considering the nature of head and tail entities contained in the relational context. 
This paper proposes a novel method that captures both connections between entities and the intrinsic nature of entities, by simultaneously aggregating \textbf{RE}lational \textbf{P}aths and c\textbf{O}ntext with a unified hie\textbf{R}archical \textbf{T}ransformer framework, namely \textbf{REPORT}.
REPORT relies solely on relation semantics and can naturally generalize to the fully-inductive setting, where KGs for training and inference have no common entities.
In the experiments,
REPORT performs consistently better than all baselines on almost all the eight version subsets of two fully-inductive datasets.
Moreover. REPORT is interpretable by providing each element's contribution to the prediction results.
\end{abstract}
\section{Introduction}
\begin{figure}[t]
\centering
\includegraphics[width=0.95\columnwidth]{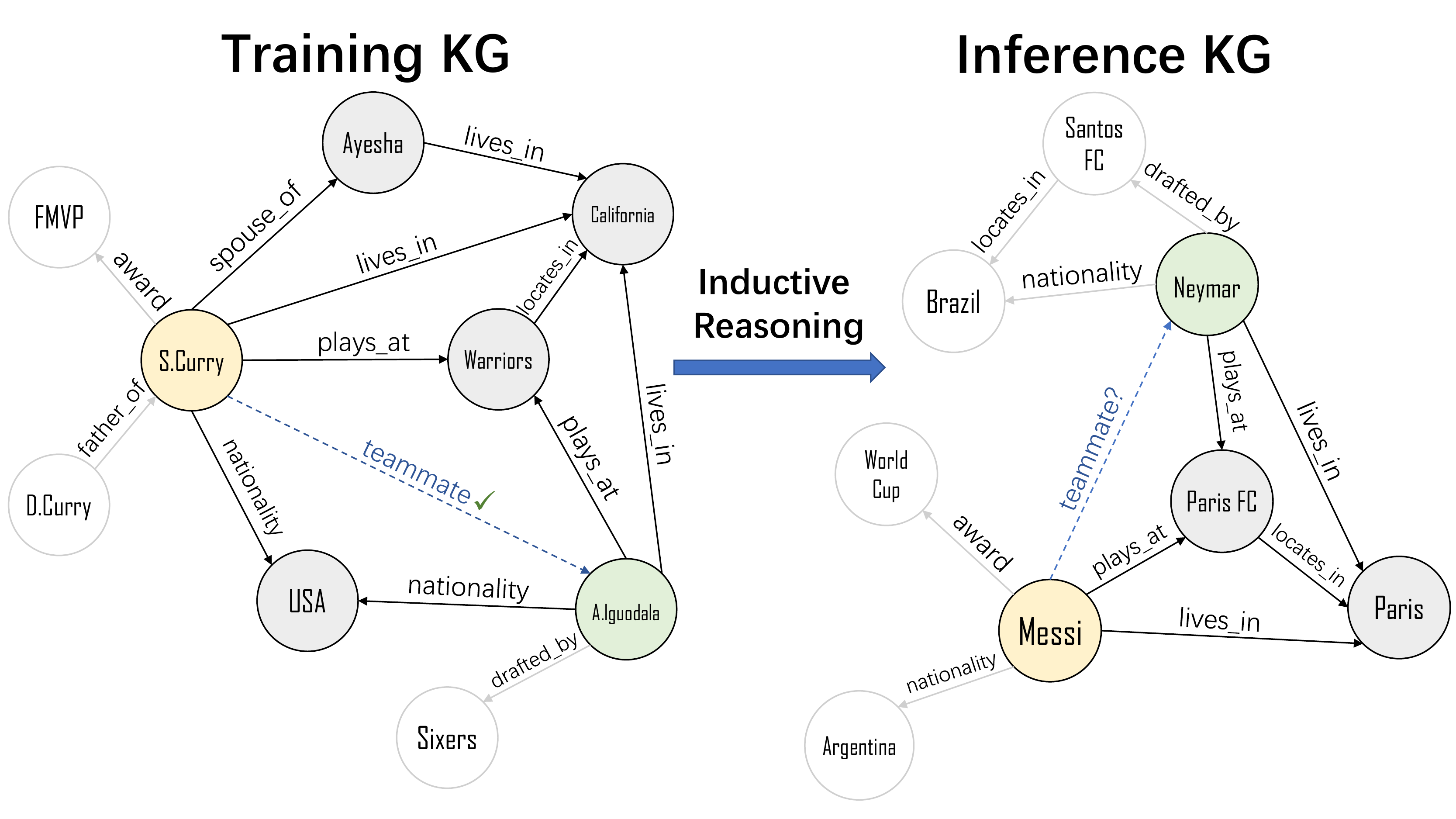} 
\caption{Inductive relation prediction which learns from a training KG and generalizes to another KG with no shared entities for inference. Yellow and green nodes denote the head and tail entities. Blue dashed line denotes the relation to be predicted.}
\vspace{-0.2in}
\label{fig1}
\end{figure}

Knowledge Graphs (KGs) are directed graphs composed of entities as nodes and relations as different types of edges. Each edge is a triplet of the form (\textit{head entity}, \textit{relation}, \textit{tail entity}), a.k.a. a \textit{fact}.
KGs, no matter how large, are usually incomplete. Various approaches have thus been proposed to predict missing links in KGs, of which the embedding-based methods became a dominant paradigm~\cite{transe,conve}. 
Most of these approaches are applicable solely to the \textit{transductive} setting, which assumes that the set of entities in a KG is fixed. 
However, in practical applications, new entities always emerge over time, e.g., new users and products on e-commerce platforms~\cite{grail}. This requires the \textit{inductive} ability to generalize to unseen entities.
Thus, inductive relation prediction, which aims to infer missing relations on KGs containing new entities, is receiving increasing attention (see Figure~\ref{fig1} for an illustration of this task).

In order to deal with potential new entities, the key to inductive relation prediction is to use information irrelevant to specific entities. 
The most representative technique of this kind is through rule mining~\cite{neurallp,rulen,drum}, which mines first-order logic rules from a given KG and uses a weighted composition of these rules for inference. Each rule could be viewed as a \textit{relational path} composed of a sequence of relations starting from a head entity to a tail entity, indicating the presence of a target relation between the two entities, e.g., the simple rule $(X, {\small\tt plays\_at}, Y) \wedge(Y, {\small\tt plays\_at}^{-1}, Z) \rightarrow (X, {\small\tt teammate}, Z)$ derived from the KG shown in Figure~\ref{fig1}.\footnote{Here we use $r^{-1}$ to denote the inverse of relation $r$.} These relational paths, which are in symbolic form and irrelevant to specific entities, are therefore inductive and highly interpretable. Some recent work also proposed to aggregate relational paths between two entities into a subgraph and predict missing relations by mining higher-order semantics contained in the topological structure of the subgraph~\cite{grail,mai2021communicative,tact,conglr,kwak2022subgraph}.

\begin{figure*}[t]
\centering
\includegraphics[width=0.85\textwidth]{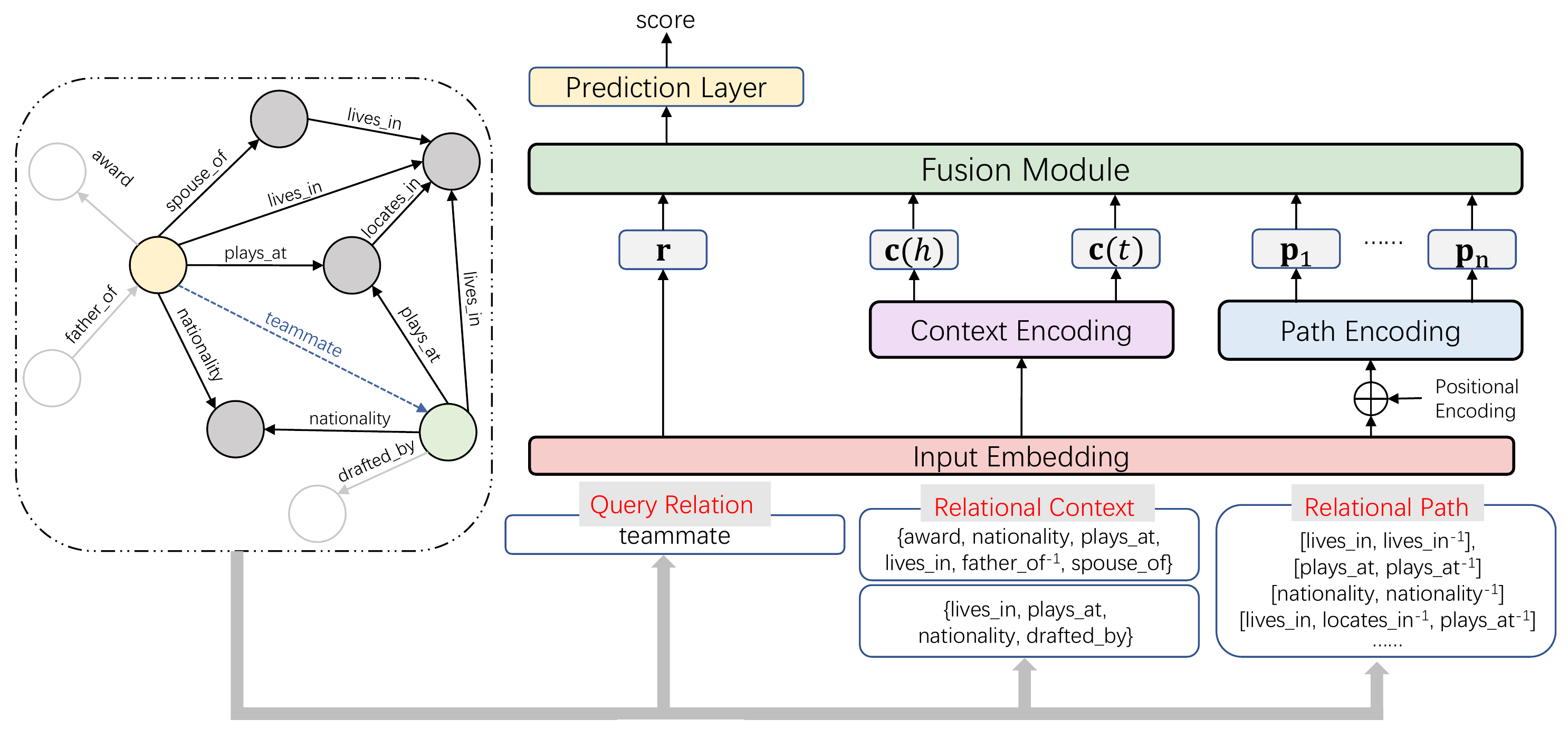} 
\caption{The overall framework of REPORT, which generates the prediction score for a query fact $(h,r,t)$ by considering (1) relational paths from head entity $h$ to tail entity $t$, (2) relational context of the two entities $h$ and $t$ respectively, and (3) the query relation $r$ itself, via a hierarchical Transformer architecture.}
\vspace{-0.2in}
\label{fig2}
\end{figure*}

Nevertheless, relational paths, no matter used individually or together as a subgraph, describe only the relationships between a pair of entities while ignoring the intrinsic attributes of the entities themselves. 
Previous research~\cite{gake,pathcon} has shown that \textit{relational context}, which is defined as a set of neighboring relations of an entity in a KG, can reflect the nature of the given entity and hence is beneficial to relation prediction. Take the entity ${\small\tt A.\ Iguodala}$ in Figure~\ref{fig1} as an example. The neighboring relations ${\small\tt lives\_in}$ and ${\small\tt nationality}$ indicate that this entity is a person, while the neighboring relations ${\small\tt plays\_at}$ and ${\small\tt drafted\_by}$ further suggest that this entity is an athlete, making it more likely to be a tail entity of the relation ${\small\tt teammate}$. Therefore, it is necessary to take relational context as a supplement to relational paths for inductive relation prediction.

In this paper, we propose \textbf{REPORT}, a new approach that aggregates \textbf{RE}lational \textbf{P}aths and c\textbf{O}ntext with hie\textbf{R}archical \textbf{T}ransformers for inductive relation prediction. The overall framework is shown in Figure~2, which consists of two path and context encoding modules at the bottom and a fusion module at the top.
All modules are constructed with Transformer blocks \cite{DBLP:conf/nips/VaswaniSPUJGKP17}.
The path encoding module encodes individual relational paths between a pair of head and tail entities, while the context encoding module encodes relational context around each of the two entities. The fusion module then aggregates path and context representations and employs an adaptively weighted combination of these components to make final predictions. 
As REPORT relies solely on relation types without entity IDs, it is naturally inductive and can handle potential new entities unseen during training. 
Moreover, it uses a weighted combination of paths and context for inference, thus is highly interpretable.

Compared with previous work that took advantage of relational paths and/or context for transductive relation prediction \cite{ptranse,gake,pathcon}, this work focuses on the fully-inductive setting and, for the first time to our knowledge, simultaneously represents and adaptively aggregates all paths and context in a unified, hierarchical Transformer architecture. 
Different from Hitter~\cite{hitter}, which encodes entities' 1-hop neighborhood for transductive relation prediction, REPORT is designed to aggregate relational context and multi-hop paths for inductive reasoning.
There were also previous works that generalized transductive KG embedding methods to the inductive setting, by initializing entity embeddings with their respective relational context~\cite{nodepiece,morse}. These approaches, however, are restrained by specific scoring functions used in transductive KG embedding and lack interpretability.

We carry out empirical evaluation on two benchmarking datasets of inductive relation prediction. 
Our approach REPORT outperforms all baselines on almost all eight versions of the two benchmarks, showing its superiority under fully-inductive setting. Ablation studies further verify the necessity of modeling both relational paths and context, and case studies verify the interpretability of our approach.

\vspace{-0.05in}
\section{Proposed Method}
\vspace{-0.05in}
\subsection{Problem Statement}
We denote a KG as $\mathcal{G=(E,R,F)}$, where $\mathcal{E}$ is the set of entities and $\mathcal{R}$ the set of relations in $\mathcal{G}$. 
$\mathcal{F}=\{(h,r,t)\subseteq \mathcal{E\times R\times E}\}$ is the set of facts appeared in $\mathcal{G}$, where $h$ and $t$ are head and tail entities and $r$ refers to a relation. 
Definitions of inductive relation prediction task, relational paths, and relational context are given below.

\textbf{Definition 1 (Inductive Relation Prediction)}
The inductive relation prediction task is learning from a training KG $\mathcal{G}_{T}=(\mathcal{E}_T,\mathcal{R}_T,\mathcal{F}_T)$, and generalizing to another KG $\mathcal{G}_{I}=(\mathcal{E}_I,\mathcal{R}_I,\mathcal{F}_I)$ for inference.
It needs to score whether a query fact $(h,r,t)$ is correct, where $(h,r,t)\in  \mathcal{E}_I\times \mathcal{R}_I\times \mathcal{E}_I$ and $(h,r,t)\notin \mathcal{F}_{I}$. 
This task is under a fully-inductive setting, where $\mathcal{G}_{T}$ and $\mathcal{G}_{I}$ have no overlapping entities ($\mathcal{E}_T\cap \mathcal{E}_I=\phi$), and all relations in $\mathcal{G}_{I}$ have already appeared during training ($\mathcal{R}_I \subseteq \mathcal{R}_T$) with the same semantics.

\textbf{Definition 2 (Relational Paths)}
A relational path is a sequence of consecutive relations taken from a path linking two entities in a KG.
We discard specific entities in the path in order to generalize to new entities. 
A relational path of length $k$ is represented in the form $[r_1,\cdots,r_k]$, e.g., $[{\small\tt plays\_at}, {\small\tt plays\_at}^{-1}]$ is a length-2 path starting from the entity ${\small\tt S.\ Curry}$ to ${\small\tt A.\ Iguodala}$ with the intermediate node ${\small\tt Warriors}$ in Figure~\ref{fig1}.

\textbf{Definition 3 (Relational Context)}
The relational context of an entity is a set of outgoing relations directly connected to that entity in a KG. Relational context of size $k$ is represented in the form $\{r_1,\cdots,r_k\}$, e.g., the relational context of ${\small\tt Warriors}$ is $\{{\small\tt play\_at},\ {\small\tt locate\_in}^{-1}\}$ in Figure~\ref{fig1}.

\begin{table*}[htbp]
\renewcommand\arraystretch{0.7}
\setlength\tabcolsep{2.4pt}
  \centering
  \small
   \begin{tabular}{cccccccccccccccccccccccc}
    \toprule
    \multirow{3}[6]{*}{Model} & \multicolumn{11}{c}{WN18RR}                                                           &       & \multicolumn{11}{c}{FB15k-237} \\
\cmidrule{2-12}  \cmidrule{14-24}        & \multicolumn{2}{c}{v1} &       & \multicolumn{2}{c}{v2} &       & \multicolumn{2}{c}{v3} &       & \multicolumn{2}{c}{v4} &       & \multicolumn{2}{c}{v1} &       & \multicolumn{2}{c}{v2} &       & \multicolumn{2}{c}{v3} &       & \multicolumn{2}{c}{v4} \\
\cmidrule{2-3}\cmidrule{5-6}\cmidrule{8-9}\cmidrule{11-12}\cmidrule{14-15}\cmidrule{17-18}\cmidrule{20-21}\cmidrule{23-24}          & H@10 & MRR   &       & H@10 & MRR   &       & H@10 & MRR   &       & H@10 & MRR   &       & H@10 & MRR   &       & H@10 & MRR   &       & H@10 & MRR   &       & H@10 & MRR \\
    \midrule
    NeuraLP & 74.37 & 71.74 &       & 68.93 & 68.54 &       & 46.18 & 44.23 &       & 67.13 & 67.14 &       & 52.92 & 46.13 &       & 58.94 & 51.85 &       & 52.90  & 48.70  &       & 55.88 & 49.54 \\
    DRUM  & 74.37 & 72.46 &       & 68.93 & 68.82 &       & 46.18 & 44.96 &       & 67.13 & 67.27 &       & 52.92 & 47.55 &       & 58.73 & 52.78 &       & 52.90  & 49.64 &       & 55.88 & 50.43 \\
    RuleN & 80.85 & 79.15 &       & 78.23 & 77.82 &       & 53.39 & 51.53 &       & 71.59 & 71.65 &       & 49.76 & 45.97 &       & 77.82 & \underline{69.08} &       & 87.69 & \textbf{73.68} &       & 85.60  & \textbf{74.19} \\
    GraIL & 82.45 & \underline{80.45} &       & 78.68 & 78.13 &       & 58.43 & \underline{54.11} &       & 73.41 & 73.84 &       & 64.15 & \underline{48.56} &       & 81.80  & 62.54 &       & 82.83 & 70.35 &       & 89.29 & 70.60 \\
    CoMPILE & 83.60  & 78.28 &       & 79.82 & \underline{79.61} &       & 60.69 & 53.97 &       & 75.49 & \underline{75.34} &       & 67.64 & 50.52 &       & 82.98 & 65.54 &       & 84.67 & 66.95 &       & 87.44 & 63.69 \\
    TACT  & 84.04 & —     &       & 81.63 & —     &       & 67.97 & —     &       & 76.56 & —     &       & 65.76 & —     &       & 83.56 & —     &       & 85.20  & —     &       & 88.69 & — \\
    RPC-IR & 85.11 & —     &       & 81.63 & —     &       & 62.40  & —     &       & 76.35 & —     &       & 67.56 & —     &       & 82.53 & —     &       & 84.36 & —     &       & 89.22 & — \\
    ConGLR & \underline{85.64} & —     &       & \textbf{92.93} & —     &       & \underline{70.74} & —     &       & \textbf{92.90}  & —     &       & \underline{68.29} & —     &       & \underline{85.98} & —     &       & \underline{88.61} & —     &       & \underline{89.31} & — \\
    \midrule
    REPORT & \textbf{88.03} & \textbf{80.95} &       & \underline{85.83} & \textbf{82.01} &       & \textbf{72.31} & \textbf{58.38} &       & \underline{81.46} & \textbf{77.43} &       & \textbf{71.69} & \textbf{53.22} &       & \textbf{88.91} & \textbf{70.62} &       & \textbf{91.62} & \underline{71.51} &       & \textbf{92.28} & \underline{71.28} \\
    \bottomrule
    \end{tabular}%

    \caption{Experiment results of inductive relation prediction on MRR and H@10. 
  NeuraLP, DRUM, and RuleN results are collected from \cite{grail}. 
  MRR of CoMPILE are obtained through our rerunning. 
  H@10 of TACT are collected from \cite{conglr}.
  Other baseline results are collected from their original literature.
  RPC-IR, ConGLR, and TACT are not compared in MRR since they do not report MRR, and there is no open code to generate such results.
  Best scores are in bold, and second-best ones are underlined.
  }
  \vspace{-0.15in}
  \label{tab:addlabel}%
\end{table*}%

\vspace{-0.05in}
\subsection{Model Overview}
\vspace{-0.05in}
Our model generates a prediction score indicating the correctness of each query fact $(h,r,t)$ given its background KG $\mathcal{G=(E,R,F)}$, where $\mathcal{G}=\mathcal{G}_T$ in the training phase and $\mathcal{G}=\mathcal{G}_I$ in the inference phase.
To do this, we take as input: (1) all relational paths from head entity $h$ to tail entity $t$ within length $k$, denoted as $\mathcal{P}_k(h,t) = \{p_i|p_i=[r^i_{1},\cdots,r^i_{k_i}],k_i\leq k\}_{i=1}^n$, where $n$ is the number of paths, $k$ a length limit, and $k_i$ the length of path $p_i$; (2) the relational context of head and tail entities $h$ and $t$, denoted as $\mathcal{C}(h)=$ $\{r^h_1, \cdots, r^h_{k_h}\}$ and $\mathcal{C}(t)=\{r^t_1, \cdots, r^t_{k_t}\}$; and (3) the query relation $r$ as well. We add an inverse fact $(t,r^{-1},h)$ into the background KG for each $(h,r,t)\in\mathcal{F}$ to get more meaningful paths and context.

Our model REPORT employs a hierarchical Transformer architecture to encode above information, and fuse them with the query relation to make the final prediction. 
Specifically, REPORT introduces two separate encoding modules at the bottom, to encode relational paths in $\mathcal{P}_k(h,t)$, and the relational context $\mathcal{C}(h)$ and $\mathcal{C}(t)$. 
These encoded paths and context are then aggregated by a fusion module at the top, and an adaptively weighted combination of which is finally used to score the query fact $(h,r,t)$. 
Since a same path/context may contribute differently to the final prediction given different query relations, we include the query relation $r$ into the fusion module and make the prediction well adapted to this relation. 


\subsection{Path Encoding Module}
The path encoding module encodes each relational path in $\mathcal{P}_k(h,t)$ and learns their representations.
To construct inputs to the path encoding module, we add a special token
${\small\tt [PCLS]}$ at the first place of each relational path, which is in the form of a relation sequence, to aggregate information.
Taking relational path $p_i$ as an example, we feed the sequence $(r^i_0,r^i_{1},\cdots,r^i_{k_i})$, where $r^i_0$ stands for ${\small\tt [PCLS]}$, into an embedding layer to output a sequence of embeddings.
For elements $\{\mathbf{ele}^{p_i}_j\}_{j=0}^{k_i}$ in the embedding sequence, we add a positional encoding to each of them to preserve the order of relations, i.e.,
\begin{equation}
    \mathbf{x}^{0}_j = \mathbf{ele}^{p_i}_j + \mathbf{pos}_{j},
\end{equation}
where $\mathbf{ele}^{p_i}_j$ is the embedding for the element $r^i_j$, $\mathbf{pos}_{j}$ is the corresponding positional embedding for the $j$-th position, and $\mathbf{x}^{0}_j$ is the input to path encoding module of $r^i_j$.

The path encoding module, a stack of $L_P$ successive Transformer layers, then encodes the inputs as:
 \begin{equation}
 \label{tf}
     \mathbf{x}^{l}_j = \mathrm{Transformer}(\mathbf{x}^{l-1}_j),\quad l=1,\cdots,L_p,
 \end{equation}
where $\mathbf{x}^{l}_j$ is the hidden state of $r^i_j$ after the $l$-th Transformer layer. 
We take $\mathbf{x}^{L_p}_0$ as the representation of $p_i$, denoted as $\mathbf{p}_i$. 
Following the above procedure, we encode all relational paths in $\mathcal{P}_k(h,t)$.

\subsection{Context Encoding Module}
The context encoding module encodes relational context $\mathcal{C}(h)$ and $\mathcal{C}(t)$, each of which is a set of relations.
We insert different tokens ${\small\tt [HCLS]}$ and ${\small\tt [TCLS]}$ into $\mathcal{C}(h)$ and $\mathcal{C}(t)$ to indicate different roles of head and tail entities, when constructing inputs.


Taking $\mathcal{C}(h)$ as an example, we feed $(r^h_0,r^h_1, \cdots, r^h_{k_h})$, where $r^h_0$ stands for ${\small\tt [HCLS]}$, into the embedding layer.
Denote $\mathbf{ele}^{h}_j$ as the embedding of $r^h_j$, whose input to the context encoding module is:
\begin{equation}
    \mathbf{y}^{0}_j = \mathbf{ele}^{h}_j.
\end{equation}
Similar to equation~\ref{tf}, 
it is encoded by the context encoding module, composed of $L_c$ successive Transformer layers, which is written as:
 \begin{equation}
 \label{tf_context}
     \mathbf{y}^{l}_j = \mathrm{Transformer}(\mathbf{y}^{l-1}_j),\quad l=1,\cdots,L_c.
 \end{equation}
We extract $\mathbf{y}^{L_c}_0$, which is the last-layer hidden state of ${\small\tt [HCLS]}$ as the context representation of $\mathcal{C}(h)$, denoted as $\mathbf{c}(h)$.
By changing ${\small\tt [HCLS]}$ to ${\small\tt [TCLS]}$ and repeat the same process, we can encode $\mathcal{C}(t)$ as $\mathbf{c}(t)$.

 \subsection{Fusion Module}
 The fusion module aggregates obtained representations uniformly and generates prediction results.
It is constructed by $L_f$ layers of Transformer encoders, which takes 
path representation $\{\mathbf{p}_i\}_{i=1}^n$, context representation $\mathbf{c}(h)$ and $\mathbf{c}(t)$ and query relation $r$'s embedding $\mathbf{r}$ as input.
We formalize them into $[\mathbf{r},\mathbf{c}(h),\mathbf{c}(t),\mathbf{p}_1,\cdots,\mathbf{p}_n]$, denoted as $[\mathbf{h}^{0}_0,\cdots,\mathbf{h}^{0}_{n+2}]$.
Then the update of each layer is:
 \begin{equation}
 \label{tf_context}
     \mathbf{h}^{l}_j = \mathrm{Transformer}(\mathbf{h}^{l-1}_j),\quad l=1,\cdots,L_f.
 \end{equation}

We take the output representation of $\mathbf{r}$ for final prediction, which is $\mathbf{h}^{L_f}_0$.
For specific query relations, $\mathbf{h}^{L_f}_0$ is supposed to adaptively combine paths and context for final prediction by attention mechanism\cite{DBLP:conf/nips/VaswaniSPUJGKP17}.
We feed it into the prediction layer, which comprises two linear transformations with GELU activation~\cite{gelu} in between and a final sigmoid normalization, which can be written as:
\begin{align}
    s=\mathrm{sigmoid}(\mathbf{W}_2(\mathrm{GELU}(\mathbf{W}_1\mathbf{h}^{L_f}_0+\mathbf{b}_1))+\mathbf{b}_2).
\end{align}
Here, weight matrix $\mathbf{W}_1,\mathbf{W}_2$ and bias $\mathbf{b}_1,\mathbf{b}_2$ are learnable parameters. $s$ is the score indicating query fact's correctness.

\begin{table*}[htb]
\renewcommand\arraystretch{0.3}
\setlength\tabcolsep{3pt}
  \centering
  \small
    \begin{tabular}{clr}
    \toprule
    \multicolumn{1}{c}{\textbf{Query Fact}} & \multicolumn{1}{c}{\textbf{Component}} & \multicolumn{1}{c}{\textbf{Score}} \\
    \midrule
    \multicolumn{1}{l}{\multirow{3}[1]{0.38\textwidth}{
    (\textcolor{my_blue}{University\ of\ Arizona}, \textcolor{magenta}{field\_of\_study}, \textcolor{my_green}{Finance})}} 
    & [people/institution$^{-1}$, people/institution, field\_of\_study] & 0.222 \\
    & [people/institution$^{-1}$, people/study, field\_of\_study$^{-1}$, field\_of\_study] & 0.188\\
    & [field\_of\_study, study/students\_majoring, field\_of\_study$^{-1}$, field\_of\_study] & 0.061\\
    \midrule

    \multicolumn{1}{l}{\multirow{3}[2]{0.3\textwidth}{(\textcolor{my_blue}{Sony BMG Music Entertainment},\newline{}\textcolor{magenta}{music/artist},\textcolor{my_green}{Christina Aguilera})}} 
    & [music/artist, music\_genere/artist$^{-1}$, music\_genere/artist] & 0.187 \\
    & \multicolumn{1}{p{27.94em}}{\textcolor{my_green}{tail:\{celebrity/dated$^{-1}$,music\_genere/artist$^{-1}$,celebrity/canoodled$^{-1}$,\newline{}people/place\_lived,award/award\_winner$^{-1}$\}}} & 0.148 \\
    & [location, vacation\_choice\_of, nominated\_with,  celebrity/canoodled] & 0.120 \\
    \midrule
    
    \multicolumn{1}{l}{\multirow{2}[2]{0.35\textwidth}{
    (\textcolor{my_blue}{Sardina}, \textcolor{magenta}{administrative\_division\_of}, \textcolor{my_green}{Italy})
    }} 
    & [administrative\_parent] & 0.368\\
    & [location\_contain$^{-1}$] & 0.326 \\
    &\textcolor{my_blue}{head:\{administrative\_parent,location\_contain$^{-1}$,vacation\_choice\_of\}} & 0.221 \\
    
    \bottomrule
    \end{tabular}%
  \label{tab:addlabel}%
  \caption{Some cases for interpretability of REPORT. Head entities, relations, and tail entities are in blue, red and green, respectively. Black components denote relational paths. Blue and green components denote the relational context of the head and tail entities, respectively.}
\end{table*}%

\begin{table}[htb]
\renewcommand\arraystretch{0.2}
\setlength\tabcolsep{1.7pt}
  \centering
  \small
    \begin{tabular}{lccccccccccc}
    \toprule
    \multirow{2}[4]{*}{} & \multicolumn{4}{c}{WN18RR}       &      & \multicolumn{4}{c}{FB15k-237} \\
\cmidrule{2-5}\cmidrule{7-10}         & v1   & v2   & v3   & v4    &      & v1   & v2   & v3   & v4    \\
    \midrule
    \textbf{REPORT} & 88.03 & 85.83 & 72.31 & 81.46  &      & 71.69 & 88.91 & 91.62 & 92.28  \\
    w\textbackslash{}o context & 83.78 & 81.63 & 63.31 & 76.35  &      & 61.22 & 79.81 & 77.86 & 72.93  \\
    w\textbackslash{}o path & 27.66 & 31.29 & 38.51 & 28.90  &      & 38.54 & 59.94 & 41.39 & 35.50  \\
    \bottomrule
    \end{tabular}%
  \label{tab:addlabel}%
  \caption{Ablation results (Hits@10). ``w\textbackslash{}o context'' and ``w\textbackslash{}o path'' mean 
  discarding context and path representations, respectively.}

\end{table}%

 \subsection{Model Training}
 Our model is trained on the training KG $\mathcal{G}_t$. 
 Given positive facts $\mathcal{F}^+\subseteq\mathcal{F}_t$ and corresponding negative facts $\mathcal{F}^-$ 
, we train our model using $\mathcal{F}^+ \cup \mathcal{F}^-$ with the binary cross-entropy loss:
 \begin{equation}
     \mathcal{L}=-\sum_{f_i \in \mathcal{F}^+ \cup \mathcal{F}^-}(y_i\log{s_i}+(1-y_i)\log{(1-s_i)}),
 \end{equation}
 where $s_i$ is the prediction score REPORT outputs for the query fact $f_i$. $y_i\in\{0,1\}$ indicates negative or positive label.
 
$\mathcal{F}^-$ are sampled by corrupting head or tail entity in each positive fact $(h,r,t)\in\mathcal{F}^+$ with a 
random entity from $\mathcal{E}_t$, i.e.,
 \begin{equation}
     \mathcal{F}^-=\{(h',r,t)\, or\, (h,r,t')\notin\mathcal{F}^+|(h,r,t)\in\mathcal{F}^+\}.
 \end{equation}
 

\section{Experiments}


\subsection{Datasets and Baseline methods}
We use two benchmark datasets generated by~\cite{grail}, for inductive relation prediction. Each contains four subsets with varying sizes sampled from WN18RR~\cite{wn18rr} and FB15k-237~\cite{fb15k-237}, respectively.

We compare REPORT with some state-of-the-art methods, which are in two types in general. 
The first is based on rules or relational paths, including RuleN~\cite{rulen}, DRUM~\cite{drum}, NeuraLP~\cite{neurallp} and RPC-IR~\cite{rpc-ir}.
The second type reasons through enclosing subgraphs, including GraIL~\cite{grail}, CoMPILE~\cite{mai2021communicative}, TACT~\cite{tact}, and ConGLR~\cite{conglr}.

\subsection{Training and Evaluation}
We set each module contain a 2-layer Transformer encoder with 4 attention heads.
The embedding size and hidden size are set to 64 and 128.
During training, experiments run for 50 epochs with early stopping, with the batch size of 128.
Grid search is also applied for:

\begin{itemize}[nosep,left=1em]
\item learning rate: \{5e-5, 1e-4, 5e-4, 1e-3, 5e-3\}
\item dropout rate: \{0.1, 0.2, 0.3, 0.4, 0.5, 0.6\}
\end{itemize}

Our baselines use rules up to 4~\cite{rulen}, or subgraphs
containing all paths with a maximum length of 4 between entities~\cite{grail}.
To be consistent with them, we extract all relational paths up to 4.
We set the upper limit of path number for each fact to 300, which covers almost all cases, or we will sample 300 relational paths at each epoch.

We apply the standard evaluation protocols proposed by~\cite{grail}.
Each fact for inference is ranked among 50 other negative facts sampled by replacing $h$ or $t$ with a random entity.
We report mean reciprocal rank (MRR) and Hits@10(H@10) averaged in 5 runs.

\subsection{Inductive Relation Prediction}
Performances of different models are shown in Table~1.
REPORT achieves better performance on most of the subsets compared to other SOTA baselines in both metrics, especially by around three absolute points over FB15k-237 in Hits@10 on average.
As for MRR, REPORT performs better or at least equally well with compared baselines over all subsets. 
It verifies our claim that uniformly aggregating relational context with relational paths could enhance inductive relation prediction results.

On WN18RR, REPORT performs closed to the strongest baseline ConGLR in Hits@10. 
After observing its subsets, we find they are sparse with few types of relations.
Thus, the relational context we use captures limited information from such KG.
Nevertheless, by a significant margin, REPORT still outperforms all other baselines.

\subsection{Case Study}
REPORT is highly interpretable and can explain its prediction results by listing the contribution scores of all elements, i.e., each relational path and context.
Since query relation is left as a special token which aggregates all elements in the fusion module,
its attention weights in the last layer indicate different elements' contributions to the result.
We take the average attention weight over all attention heads as the contribution scores, which sum to 1 for all elements of one query fact.
Table~2 shows the top 3 elements with the highest contribution scores in several cases.
Correlations between these elements and query facts are easy to understand.
REPORT aggregates this information with adaptive weights thus can perform well in reasoning.

\subsection{Ablation Study}
We conduct ablation studies on relational paths and context, by discarding corresponding representations in the fusion module.
That means inputs to the fusion module are $[\mathbf{r},\mathbf{c}(h),\mathbf{c}(t)]$ and $[\mathbf{r},\mathbf{p}_1,\cdots,\mathbf{p}_n]$, respectively.
Results are shown in table~3.

Results demonstrate the importance of both kinds of information for prediction,
validating that relational context is an important supplement. 
Although these two information cannot achieve satisfactory results when used alone, REPORT can adaptively combine their semantics to achieve better performance, verifying its ability to aggregate relational paths and context for inductive reasoning.

\section{Conclusions}
In this paper, we propose REPORT, a novel method for inductive relation prediction.
It fully uses relational paths and context, 
and aggregates them adaptively within a unified hierarchical Transformer architecture to generate prediction results.
Experimental results demonstrate the effectiveness and superiority of our method.
It can also provide explanations of the prediction result.
\bibliographystyle{IEEEbib}
\bibliography{refs}

\begin{thebibliography}{10}

\bibitem{transe}
Antoine Bordes, Nicolas Usunier, Alberto Garcia-Dur\'{a}n, Jason Weston, and
  Oksana Yakhnenko,
\newblock ``Translating embeddings for modeling multi-relational data,''
\newblock in {\em Proceedings of the 26th International Conference on Neural
  Information Processing Systems - Volume 2}, Red Hook, NY, USA, 2013, NIPS'13,
  p. 2787–2795, Curran Associates Inc.

\bibitem{conve}
Tim Dettmers, Pasquale Minervini, Pontus Stenetorp, and Sebastian Riedel,
\newblock ``Convolutional 2d knowledge graph embeddings,''
\newblock in {\em Proceedings of the AAAI conference on artificial
  intelligence}, 2018, vol.~32.

\bibitem{grail}
Komal Teru, Etienne Denis, and Will Hamilton,
\newblock ``Inductive relation prediction by subgraph reasoning,''
\newblock in {\em International Conference on Machine Learning}. PMLR, 2020,
  pp. 9448--9457.

\bibitem{neurallp}
Fan Yang, Zhilin Yang, and William~W Cohen,
\newblock ``Differentiable learning of logical rules for knowledge base
  reasoning,''
\newblock {\em Advances in neural information processing systems}, vol. 30,
  2017.

\bibitem{rulen}
Christian Meilicke, Manuel Fink, Yanjie Wang, Daniel Ruffinelli, Rainer
  Gemulla, and Heiner Stuckenschmidt,
\newblock ``Fine-grained evaluation of rule-and embedding-based systems for
  knowledge graph completion,''
\newblock in {\em International semantic web conference}. Springer, 2018, pp.
  3--20.

\bibitem{drum}
Ali Sadeghian, Mohammadreza Armandpour, Patrick Ding, and Daisy~Zhe Wang,
\newblock ``Drum: End-to-end differentiable rule mining on knowledge graphs,''
\newblock {\em Advances in Neural Information Processing Systems}, vol. 32,
  2019.

\bibitem{mai2021communicative}
Sijie Mai, Shuangjia Zheng, Yuedong Yang, and Haifeng Hu,
\newblock ``Communicative message passing for inductive relation reasoning.,''
\newblock in {\em AAAI}, 2021, pp. 4294--4302.

\bibitem{tact}
Jiajun Chen, Huarui He, Feng Wu, and Jie Wang,
\newblock ``Topology-aware correlations between relations for inductive link
  prediction in knowledge graphs,''
\newblock in {\em Proceedings of the AAAI Conference on Artificial
  Intelligence}, 2021, vol.~35, pp. 6271--6278.

\bibitem{conglr}
Qika Lin, Jun Liu, Fangzhi Xu, Yudai Pan, Yifan Zhu, Lingling Zhang, and
  Tianzhe Zhao,
\newblock ``Incorporating context graph with logical reasoning for inductive
  relation prediction,''
\newblock in {\em Proceedings of the 45th International ACM SIGIR Conference on
  Research and Development in Information Retrieval}, 2022, pp. 893--903.

\bibitem{kwak2022subgraph}
Heeyoung Kwak and Hyunkyung Bae~Kyomin Jung,
\newblock ``Subgraph representation learning with hard negative samples for
  inductive link prediction,''
\newblock in {\em ICASSP 2022-2022 IEEE International Conference on Acoustics,
  Speech and Signal Processing (ICASSP)}. IEEE, 2022, pp. 4768--4772.

\bibitem{gake}
Jun Feng, Minlie Huang, Yang Yang, and Xiaoyan Zhu,
\newblock ``Gake: Graph aware knowledge embedding,''
\newblock in {\em Proceedings of COLING 2016, the 26th International Conference
  on Computational Linguistics: Technical Papers}, 2016, pp. 641--651.

\bibitem{pathcon}
Hongwei Wang, Hongyu Ren, and Jure Leskovec,
\newblock ``Relational message passing for knowledge graph completion,''
\newblock in {\em Proceedings of the 27th ACM SIGKDD Conference on Knowledge
  Discovery \& Data Mining}, 2021, pp. 1697--1707.

\bibitem{DBLP:conf/nips/VaswaniSPUJGKP17}
Ashish Vaswani, Noam Shazeer, Niki Parmar, Jakob Uszkoreit, Llion Jones,
  Aidan~N. Gomez, Lukasz Kaiser, and Illia Polosukhin,
\newblock ``Attention is all you need,''
\newblock in {\em Advances in Neural Information Processing Systems 30: Annual
  Conference on Neural Information Processing Systems 2017, December 4-9, 2017,
  Long Beach, CA, {USA}}, Isabelle Guyon, Ulrike von Luxburg, Samy Bengio,
  Hanna~M. Wallach, Rob Fergus, S.~V.~N. Vishwanathan, and Roman Garnett, Eds.,
  2017, pp. 5998--6008.

\bibitem{ptranse}
Yankai Lin, Zhiyuan Liu, Huanbo Luan, Maosong Sun, Siwei Rao, and Song Liu,
\newblock ``Modeling relation paths for representation learning of knowledge
  bases,''
\newblock in {\em Proceedings of the 2015 Conference on Empirical Methods in
  Natural Language Processing}, 2015, pp. 705--714.

\bibitem{hitter}
Sanxing Chen, Xiaodong Liu, Jianfeng Gao, Jian Jiao, Ruofei Zhang, and Yangfeng
  Ji,
\newblock ``Hitter: Hierarchical transformers for knowledge graph embeddings,''
\newblock in {\em Proceedings of the 2021 Conference on Empirical Methods in
  Natural Language Processing}, 2021, pp. 10395--10407.

\bibitem{nodepiece}
Mikhail Galkin, Etienne Denis, Jiapeng Wu, and William~L. Hamilton,
\newblock ``Nodepiece: Compositional and parameter-efficient representations of
  large knowledge graphs,''
\newblock in {\em International Conference on Learning Representations}, 2022.

\bibitem{morse}
Mingyang Chen, Wen Zhang, Yushan Zhu, Hongting Zhou, Zonggang Yuan, Changliang
  Xu, and Huajun Chen,
\newblock ``Meta-knowledge transfer for inductive knowledge graph embedding,''
\newblock in {\em Proceedings of the 45th International ACM SIGIR Conference on
  Research and Development in Information Retrieval}, 2022, pp. 927--937.

\bibitem{gelu}
Dan Hendrycks and Kevin Gimpel,
\newblock ``Gaussian error linear units (gelus),''
\newblock {\em arXiv preprint arXiv:1606.08415}, 2016.

\bibitem{wn18rr}
Tim Dettmers, Pasquale Minervini, Pontus Stenetorp, and Sebastian Riedel,
\newblock ``Convolutional 2d knowledge graph embeddings,''
\newblock in {\em Proceedings of the AAAI conference on artificial
  intelligence}, 2018, vol.~32.

\bibitem{fb15k-237}
Kristina Toutanova, Danqi Chen, Patrick Pantel, Hoifung Poon, Pallavi
  Choudhury, and Michael Gamon,
\newblock ``Representing text for joint embedding of text and knowledge
  bases,''
\newblock in {\em Proceedings of the 2015 conference on empirical methods in
  natural language processing}, 2015, pp. 1499--1509.

\bibitem{rpc-ir}
Yudai Pan, Jun Liu, Lingling Zhang, Xin Hu, Tianzhe Zhao, and Qika Lin,
\newblock ``Learning first-order rules with relational path contrast for
  inductive relation reasoning,'' 2021.

\end{thebibliography}

\end{document}